\documentclass[10pt,twocolumn,letterpaper]{article}

\usepackage{iccv}              

\definecolor{iccvblue}{rgb}{0.21,0.49,0.74}
\usepackage{booktabs} 
\usepackage{multirow}
\usepackage{bbding}
\usepackage{flushend}
\usepackage[ruled]{algorithm2e}

\usepackage{float}
\usepackage{setspace}
\usepackage[pagebackref,breaklinks,colorlinks,allcolors=iccvblue]{hyperref}

\def\paperID{3239} 
\def\confName{ICCV}
\def\confYear{2025}

\title{Estimating 2D Camera Motion with Hybrid Motion Basis}

\author{
Haipeng Li$^{1}$\thanks{Equal contribution.} \hspace{0.5cm}
Tianhao Zhou$^{1*}$ \hspace{0.5cm}
Zhanglei Yang$^{1}$ \hspace{0.5cm}
Yi Wu$^{2}$ \hspace{0.5cm}
Yan Chen$^{2}$ \hspace{0.5cm}\\
Zijing Mao$^{2}$ \hspace{0.5cm}
Shen Cheng$^{3}$ \hspace{0.5cm}
Bing Zeng$^{1}$ \hspace{0.5cm}
Shuaicheng Liu$^{1}$\thanks{Corresponding author (liushuaicheng@uestc.edu.cn).} \hspace{0.5cm}
 \\
\textsuperscript{1}University of Electronic Science and Technology of China\\
\textsuperscript{2}Xiaomi Corporation \textsuperscript{3}Dexmal \\
}

\begin{document}
\maketitle

\begin{abstract}
Estimating 2D camera motion is a fundamental computer vision task that models the projection of 3D camera movements onto the 2D image plane.
Current methods rely on either homography-based approaches, limited to planar scenes, or meshflow techniques that use grid-based local homographies but struggle with complex non-linear transformations.
%
%
We introduce \textbf{CamFlow}, a novel framework that represents camera motion using hybrid motion bases: physical bases derived from camera geometry and stochastic bases for complex scenarios.
Our approach includes a hybrid probabilistic loss function based on the Laplace distribution that enhances training robustness.
For evaluation, we create a new benchmark by masking dynamic objects in existing optical flow datasets to isolate pure camera motion.
Experiments show CamFlow outperforms state-of-the-art methods across diverse scenarios, demonstrating superior robustness and generalization in zero-shot settings.
Code and datasets are available at our project page: \url{https://lhaippp.github.io/CamFlow/}.
\end{abstract}

\begin{figure}
    \centering
    \includegraphics[width=1\linewidth]{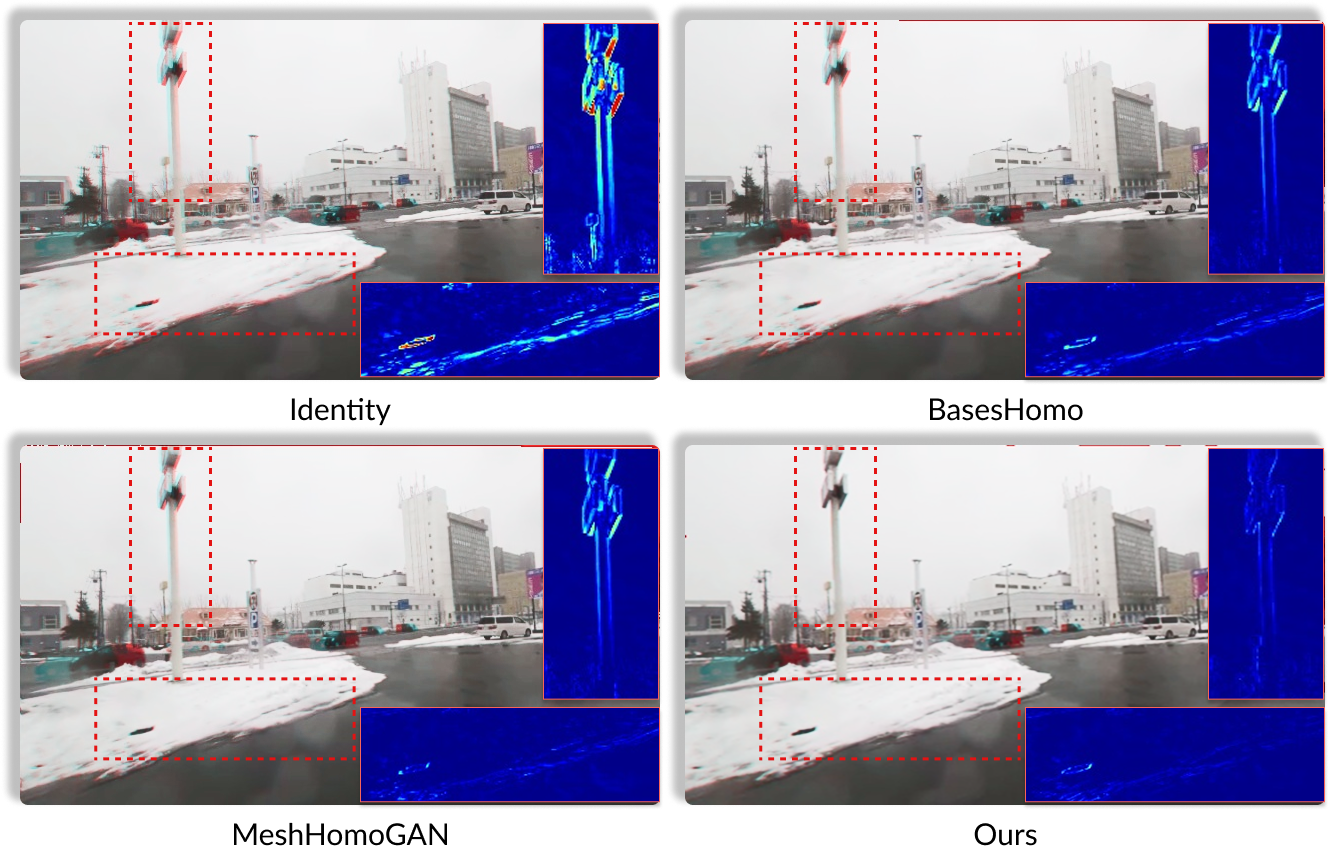}
    \caption{Comparison of camera motion estimation approaches in multi-plane scenes. Visualizations show warped source images overlaid on targets, with brighter areas in heatmaps indicating higher error. BasesHomo~\cite{BasesHomo2021} captures only background motion, while MeshHomoGAN~\cite{liu2024unsupervised} improves accuracy through local grid-based homographies. Our CamFlow, with hybrid motion basis, achieves superior representation of complex camera motion.}
    \label{fig:Teaser}
    \vspace{-1em}
\end{figure}    
\section{Introduction}
\label{intro}
%
Estimating 2D camera motion, which involves recovering the projection of 3D rotation and translation onto 2D planes~\cite{DLT}, is a cornerstone of computer vision. Given a 3D rotation matrix \(\mathbf{R}\) and translation vector \(\mathbf{t}\), the camera motion $\mathbf{M}$ can be expressed as:  
\begin{equation}  
\mathbf{M} = \mathbf{K} \left( \mathbf{R} + \mathbf{t} \frac{\mathbf{n}^T}{d} \right) \mathbf{K}^{-1},  
\label{eq:camera_pose_to_homo}  
\end{equation}  
where \(\mathbf{n}^T\) is the transpose of the normal vector to planes in the scene, \(d\) denotes the distance from the camera center to each plane, and \(\mathbf{K}\) represents the camera intrinsic matrix. 
The resulting 2D camera motion is inherently {non-linear} due to its dependence on scene depth and plane geometry. In real-world images, scenes typically consist of multiple depths and planes. As a result, different regions of an image undergo distinct transformations, leading to complex, non-linear motion patterns.
This task is essential for various computational imaging applications, such as digital video stabilization~\cite{liu2013bundled}, where accurate representation of camera motion directly enhances performance. 
Existing methods for modeling camera motion primarily fall into two categories: homography and meshflow-based approaches.

Homography, a perspective transformation, aligns two views of a planar or nearly-planar 3D scene~\cite{DLT}. 
Traditional methods typically rely on artificially extracted and matched keypoints~\cite{sift}, excelling under standard conditions but often struggling in adverse scenarios (\eg, rain, snow, and low light) or when confronted with non-planar motion caused by depth parallax or dynamic objects. 
Recent deep learning approaches mitigate the reliance on keypoints, enabling direct estimation of homography from image pairs through robust, data-driven networks. 
A notable benchmark work, BasesHomo~\cite{BasesHomo2021}, reformulates the task as learning a linear combination of an 8-dimensional motion basis, pioneering a new direction in motion estimation. 
Nevertheless, they are limited because: {\em a homography can only align a single plane\/}.

To address this limitation, MeshFlow~\cite{LiuTYSZ16} partitions the image into $N\! \times\! N$ grids, estimating a local homography for each cell and smoothing them to model non-linear camera motion. 
This approach performs well in scenes with multiple small-baseline depth variations and has become a popular choice for digital video stabilization~\cite{liu2013bundled,LiuTYSZ16}. 
Furthermore, deep meshflow variants~\cite{BasesHomo_Pami} demonstrate enhanced robustness under challenging conditions. 
However, a key limitation persists in current camera motion representation: increasing the number of grids enhances the ability to model non-linearity but raises optimization challenges~\cite{ye2019deepmeshflow}.

In this work, we introduce \textbf{CamFlow}, a novel representation that models complex camera motion through hybrid motion bases. Our key insight is that the flow field resulting from the superposition of multiple homographies is inherently non-linear (as visualized in Fig.~\ref{fig:Motion-Basis}), enabling more sophisticated motion modeling beyond single-plane limitations. Building on this observation, we establish a comprehensive hybrid basis subspace comprising:
\begin{itemize}
    \item \textbf{Physical Motion Bases}: Derived from Taylor expansion of homographic transformations up to second-order terms, our 12 physical bases model fundamental geometric transformations (rotation, translation, scaling, and perspective) that capture essential camera motion patterns;
    \item \textbf{Noisy Motion Bases}: To model complex residual motion, we construct $K$ orthogonal components through SVD decomposition of randomly sampled homographies from a Gaussian distribution, effectively complementing the physical bases by capturing higher-order motion patterns.
\end{itemize}
To stabilize the training procedure and simplify complex loss designs, we propose a hybrid probabilistic loss function that assumes motion models follow a Laplace distribution~\cite{PDC-Net}, facilitating robust and efficient optimization. 
%
%

In summary, CamFlow effectively represents complex 2D camera motion as illustrated in Fig.~\ref{fig:Teaser}, capturing both background and foreground elements while accurately modeling depth-varying motion patterns that conventional approaches struggle to represent.
To rigorously evaluate its performance, we introduce \textbf{GHOF-Cam}, a novel benchmark specifically designed for camera motion estimation by systematically masking dynamic objects and ill-posed occlusion regions in established optical flow datasets~\cite{gyroflow+}, thereby isolating pure camera-induced motion.
Through comprehensive experiments across diverse datasets under both standard and challenging conditions, we demonstrate that CamFlow consistently outperforms state-of-the-art single-plane homography and multi-planar methods in both sparse and dense camera motion estimation tasks, exhibiting superior robustness and generalization capability in real-world scenarios.
%
%
Our main contributions are:
\begin{itemize}
    \item A new hybrid motion representation that learns to model complex non-linear 2D camera motion through physically interpretable and stochastic motion bases.
    \item A novel probabilistic loss formulation based on Laplace distribution that simplifies training and stabilizes optimization without complex loss designs.
    \item A comprehensive benchmark for evaluating camera motion learning across diverse conditions. Experimental results confirm our approach's effectiveness, robustness, and generalization ability across real-world scenarios.
\end{itemize}

\vspace{-0.5em}
\section{Related works}
\label{subsec:relate_work}

\begin{figure*}[t]
    \centering
    \includegraphics[width=1\linewidth]{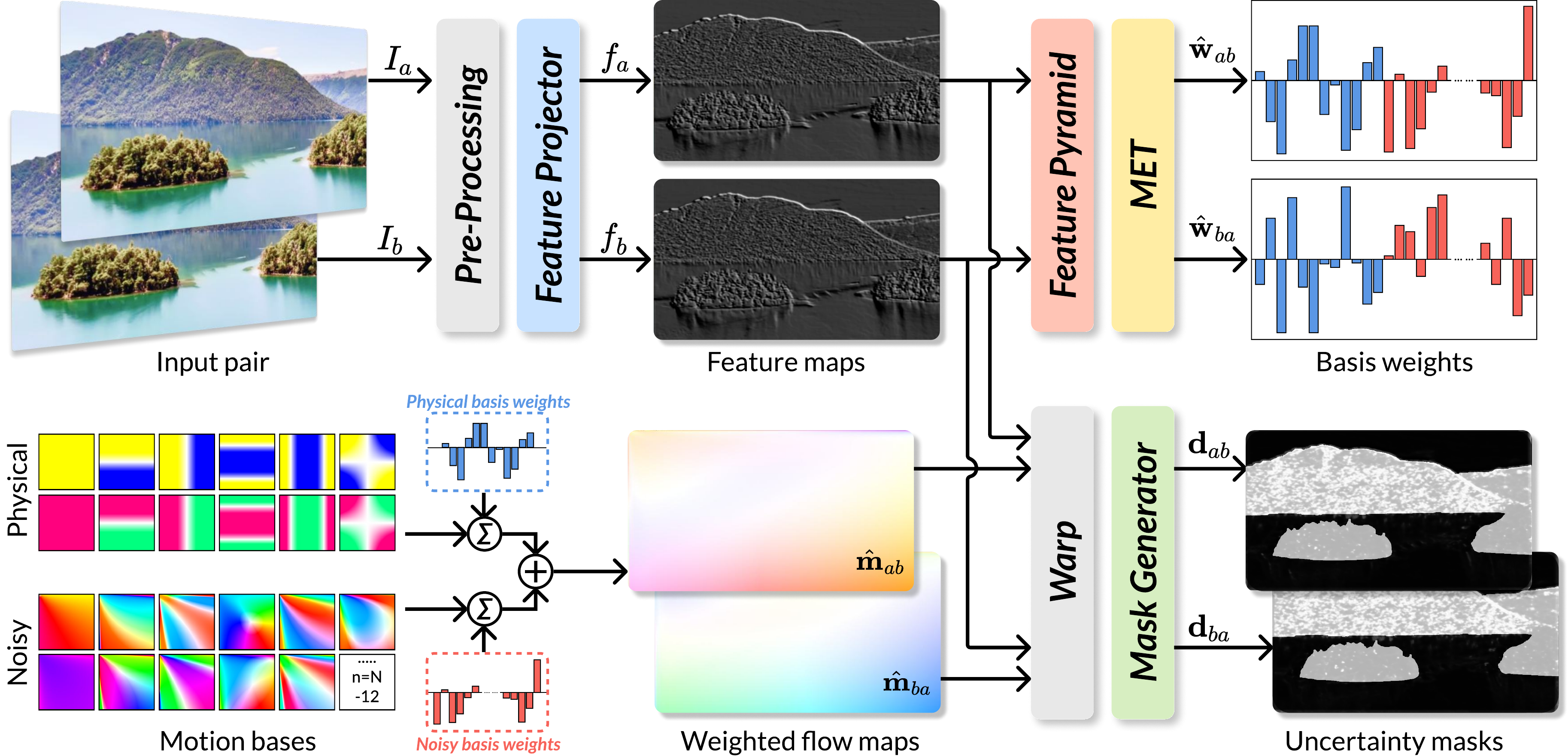}
    \vspace{-0.5em}
    \caption{Our proposed motion estimation framework. Given image pair \((I_a, I_b)\), features are extracted through a multi-scale pyramid and processed by the motion estimation transformer (MET) to compute weights for physical (blue) and noisy (red) motion bases. These weights linearly combine predefined motion bases to generate flow maps for warping. A mask generator predicts uncertainty masks \(\mathbf{d}_{ab}\) and \(\mathbf{d}_{ba}\) to reject unreliable regions, enhancing estimation robustness.}
    \label{fig:network-ppl}
\end{figure*}

\vspace{-0.3em}
\subsection{Homography Methods}
The traditional homography estimation follows three stages: feature detection (e.g., SIFT~\cite{sift}, ORB~\cite{orb}), correspondence matching~\cite{cunningham2021k}, and outlier rejection (e.g., RANSAC~\cite{ransac}, MAGSAC++~\cite{barath2020magsac++}). Learning-based methods like LIFT~\cite{lift}, SuperPoint~\cite{superpoint}, and SOSNet~\cite{sosnet} improve robustness. Optimization-based approaches~\cite{Clkn,DPCP} iteratively refine homography estimates, while deep learning methods span supervised~\cite{supervised2016,Dynamic-supervised2020,Crossresolution-supervised2021,Iterative-supervised2022,jiang2023supervised,li2024dmhomo} and unsupervised~\cite{unsupervised2018,unsupervised2020,CA-Unsupervised2020,BasesHomo2021,HomoGAN2022} frameworks. Unsupervised models have proven effective for real-world scenarios. Notable examples include CAHomo~\cite{CA-Unsupervised2020} and BasesHomo~\cite{BasesHomo2021}, which enhance feature extraction and motion constraints. HomoGAN~\cite{HomoGAN2022} integrates GAN loss~\cite{goodfellow2014generative} and Transformers~\cite{liu2021swin} for coarse-to-fine refinement. Despite this progress, homography remains a single-plane model, limiting its effectiveness in complex motion. In this paper, we introduce a hybrid motion-basis representation to model multi-plane, non-linear motion more effectively.

\vspace{-0.3em}
\subsection{Mesh-based Methods}
Mesh-based warping estimates local homographies per mesh cell, including dual-plane approaches~\cite{gao2011constructing}, patch-wise mixtures of homography~\cite{grundmann2012calibration}, flexible warping techniques like APAP~\cite{zaragoza2013projective}, Bundled Paths~\cite{liu2013bundled}, grid-based methods~\cite{LiuTYSZ16}, and cascade residual homography~\cite{shen2020ransac}. Deep learning variants include deepMeshFlow~\cite{ye2019deepmeshflow}, MeshCAHomo~\cite{liu2022content}, which merges multi-resolution meshes, and BasesMesh~\cite{liu2023unsupervised}, which applies motion bases per grid. MeshHomoGAN~\cite{liu2024unsupervised} incorporates a planarity-aware mechanism for local homography estimation. However, these methods still assume that local regions fit the homography relationship, limiting their ability to represent complex motion. To address this limitation, we propose a novel motion representation that combines multiple non-linear motion bases, enhancing performance.
\section{Method}


\subsection{Motion Basis}
\label{subsec:motion_basis}
In this work, we propose a novel motion representation through motion basis learning with deep networks to better represent the camera motion via Physical and Noisy bases.

\noindent\textbf{Physical Basis.} Consider a pixel with homogeneous coordinates $\mathbf{P}(x, y) = [x, y, 1]^T \!\in\! \mathbb{R}^3$. The physical motion $m = [\Delta x, \Delta y, 1]^T$ induced by a homography $\mathbf{H}$ is defined:
\begin{equation}
m = {\mathbf{H} \cdot \mathbf{P}(x, y)} - \mathbf{P}(x, y),
\end{equation}
After normalization, the two-dimensional coordinates (x-axis and y-axis) of the motion become:
\begin{equation}
\begin{split}
\Delta x &= \frac{h_1 x + h_2 y + h_3}{h_7 x + h_8 y + 1} - x, \\
\Delta y &= \frac{h_4 x + h_5 y + h_6}{h_7 x + h_8 y + 1} - y,
\end{split}
\end{equation}
where \( h_1, \dots, h_8 \) denote the elements of matrix \(\mathbf{H}\), and \( h_9 \) is constrained to 1.
By applying a Taylor expansion, this motion can be mapped to another subspace. For instance, expanding $\Delta x$ around the point $(x, y) =(0, 0)$ gives:
\begin{align}
\Delta x &= \frac{(h_1 - 1) x + h_2 y + h_3 - h_7 x^2 - h_8 x y}{h_7 x + h_8 y + 1}
\\
&\approx w_1 \cdot 1 + w_2 \cdot x +w_{3} \cdot y + w_4 \cdot xy \\
& \quad  + w_5 \cdot x^{2} + w_6  \cdot y^{2} + \Delta , 
\end{align}
where $w_i,i\in[1,6]$ are coefficients, the basis functions are $b=[1, x, y, xy, x^2, y^2]$ and $\Delta$ denotes the higher-order infinitesimal.
Similarly, $\Delta y$ can be decomposed into this subspace. By combining the decompositions of both $\Delta x, \Delta y$, the motion space can be represented using the 12 bases: 
\begin{equation}
    \mathbf{F}=\left\{\left(b_i, 0\right) \mid b_i \in b\right\} \cup\left\{\left(0, b_i\right) \mid b_i \in b\right\},
\end{equation}
where $b_i$ denotes the $i$-th element in $b$. Each basis can be transformed into an optical flow according to the image coordinate, as shown in Fig.~\ref{fig:network-ppl} (Physical motion bases).

\begin{figure}[t]
    \centering
    \includegraphics[width=1\linewidth]{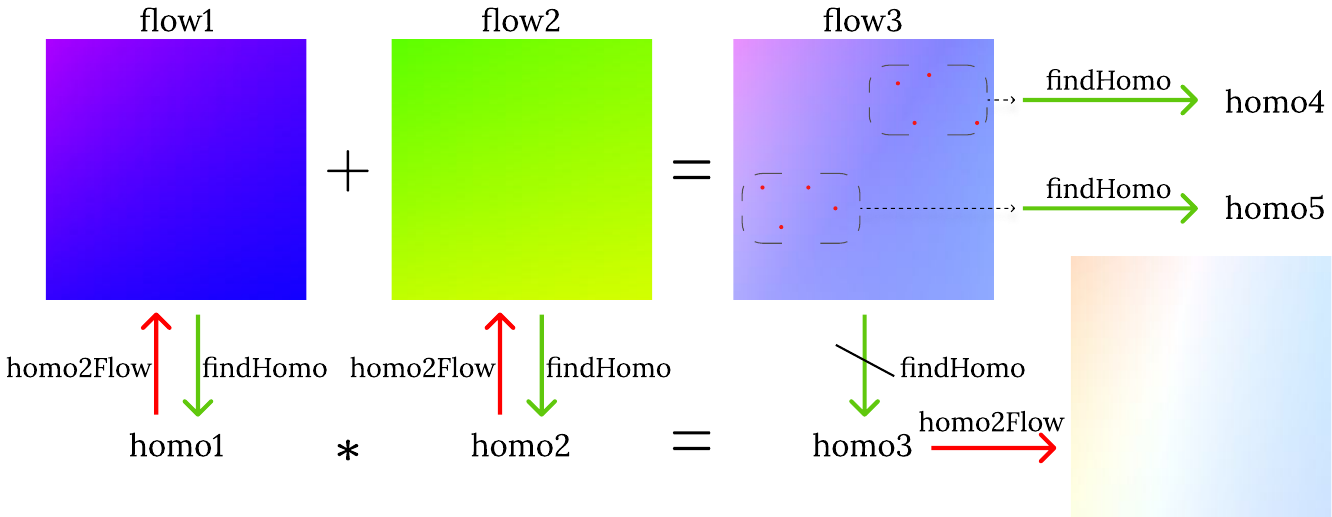}
    \vspace{-1em}
    \vspace{-1em}
    \caption{Non-linearity of flow addition. Two homography matrices generate flow1 and flow2. Adding these flows (flow3) differs from the flow derived by multiplying the original homography matrices (homo3). When sampling points from flow3 to solve for a homography, we get inconsistent solutions, proving that combined flow fields cannot be represented by a single homography.}
    \label{fig:Motion-Basis}
    \vspace{-1em}
\end{figure}

\noindent\textbf{Non-Linearity of Adding Bases.} A key finding is that combining flow fields derived from different homographies produces a non-linear motion field that cannot be represented by any single homography. This challenges the linear basis assumption in prior work~\cite{BasesHomo2021}.
As shown in Fig.~\ref{fig:Motion-Basis}, when two homography-derived flows (flow1 and flow2) are added together, the resulting flow (flow3) differs from the flow derived from multiplying the original homography matrices (homo3). Additionally, attempting to solve for a homography from points sampled on flow3 yields inconsistent solutions, confirming that the combined flow cannot be represented by any single homography.
This observation motivates our approach, which leverages more bases to model complex camera motion.

\noindent\textbf{Stochastic Basis.} While the physical bases effectively capture fundamental motion patterns, the complete space of camera motions is infinitely dimensional when considering higher-order Taylor expansions. Exhaustively modeling all possible bases becomes computationally intractable. 
To address this challenge, we leverage the expressive power of random sampling to complement our physical bases. Specifically, we generate $K$ random $3\!\times\!3$ matrices \(\left\{ \mathbf{H}^{(k)} \right\}_{k=1}^K \subseteq \mathbb{R}^{3 \times 3} \), where each matrix is formulated as:
\begin{equation}
\small
\mathbf{H} = \{ h_i \}_{i=1}^9, \text{where} \quad 
h_i = \begin{cases}
\epsilon_i \sim \mathcal{N}(0,1), & 1 \le i \le 8, \\
1, & i = 9.
\end{cases}
\end{equation}

Following BasesHomo~\cite{BasesHomo2021}, we convert random matrices into flows and apply singular value decomposition (SVD) to extract principal components. This process yields $N - 12$ stochastic bases that capture diverse motion patterns beyond physical bases, as illustrated in Fig.~\ref{fig:network-ppl} (Noisy motion bases). These stochastic bases, combined with the 12 physical basis vectors described earlier, form a comprehensive set of $N$ motion bases that enhances our ability to model complex non-linear camera motions.

\subsection{Network Structure}
\label{subsec:HEM}


The network is illustrated in Fig.~\ref{fig:network-ppl}. Following practices from previous work~\cite{HomoGAN2022,CA-Unsupervised2020}, we process video frames by: (1) random cropping to $320\! \times\! 576$ patches, (2) converting to grayscale, (3) projecting into a shallow feature space to handle luminance variations, and (4) generating a $3$-layer feature pyramid for multi-scale processing. Then we propose a Motion Estimation Transformer (MET). The MET employs a specialized architecture to separately predict weights for both physical bases (12 weights) and stochastic bases ($N-12$ weights), yielding bidirectional weights $\hat{\mathbf{w}}_{ab}$ and $\hat{\mathbf{w}}_{ba}$.
The final camera motion is computed through a linear combination of the predicted weights and their corresponding motion bases, producing bidirectional dense motion fields $\hat{\mathbf{m}}_{ab}$ and $\hat{\mathbf{m}}_{ba}$.

It is noteworthy that the confidence masks $\mathbf{d}_{ab}$ and $\mathbf{d}_{ba}$ are also crucial in predicting camera motion, as they effectively filter out dynamic objects.
First, we apply the predicted $\hat{\mathbf{m}}_{ab}$ and $\hat{\mathbf{m}}_{ba}$ to $f_a$ and $f_b$, respectively, obtaining warped feature maps $f_a^{'}$ and $f_b^{'}$.
The mask network $\mathcal{M}$~\cite{GyroFlow} inputs the concatenated features to generate weighted maps that highlight well-aligned regions:
\begin{equation}
    \mathbf{d}_{ab} = \mathcal{M}([f_a, f_b^{'}]), \mathbf{d}_{ba} = \mathcal{M}([f_b, f_a^{'}]).
\end{equation}
%

\subsection{Loss Function}
\label{subsec:loss}

\noindent\textbf{Probabilistic Motion Modeling.} Camera motion estimation faces a fundamental challenge: distinguishing between camera-induced and object motion in the scene. We therefore formulate our approach as a probabilistic model that explicitly accounts for uncertainty in motion estimation.

Building on previous findings that 2D motion follows a Laplace distribution~\cite{ilg2018uncertainty, gast2018lightweight, PDC-Net}, we model the conditional probability of the true camera motion given our prediction:
\begin{equation}
p(\mathbf{m}_{ab} \mid \hat{\mathbf{m}}_{ab};\mathbf{d}_{ab}),
\end{equation}
where $\hat{\mathbf{m}}_{ab}$ is our predicted motion field and $\mathbf{d}_{ab}$ represents our confidence in each pixel's motion estimate. Higher confidence values indicate pixels likely following camera motion, while lower values suggest pixels belonging to independently moving objects.

\noindent\textbf{Laplace Distribution Model.} We model the probability density using two conditionally independent Laplace distributions for the horizontal and vertical components:
\begin{equation}
\label{eq:nll}
\small
\begin{split}
\mathcal{L}\left(\mathbf{m}_\mathit{ab} \mid \hat{\mathbf{m}}_\mathit{ab}; \mathbf{d}_\mathit{ab}\right) &= \prod\left(\frac{1}{\sqrt{2 \sigma^2}} e^{-\sqrt{\frac{2}{\sigma^2}}\left|u-\mu_u\right|} \right. \\
&\left. \cdot \frac{1}{\sqrt{2 \sigma^2}} e^{-\sqrt{\frac{2}{\sigma^2}}\left|v-\mu_v\right|}\right),
\end{split}
\end{equation}
where $(u, v)$ are the ground-truth motion components, $(\mu_u, \mu_v)$ are the predicted motion components, and $\sigma^2$ is derived from our confidence mask $\mathbf{d}_{ab}$. This formulation allows our model to express both its prediction and its uncertainty about that prediction.

\noindent\textbf{Hybrid Loss Strategy.} Another challenge in camera motion estimation is the scarcity of ground-truth labels. To overcome this, we introduce a hybrid loss strategy with two components:
1) {Motion supervision loss} ($\ell_{\mathit{NLL}_m}$): we generate pseudo-labels using existing methods and apply negative log-likelihood (NLL) loss in both forward and backward directions:
\begin{equation}
    \footnotesize
\ell_{\mathit{NLL}_m} = -\log p(\mathbf{m}_{ab} \mid \hat{\mathbf{m}}_{ab};\mathbf{d}_{ab}) - \log p(\mathbf{m}_{ba} \mid \hat{\mathbf{m}}_{ba};\mathbf{d}_{ba}).
\end{equation}

\noindent2) {Photometric loss} ($\ell_{\mathit{NLL}_{p}}$): We apply the same probabilistic framework to enforce consistency between warped features. Given image features $f_a$ and $f_b$, we use our predicted motion to produce warped features $f_a'$ and $f_b'$:
\begin{equation}
\small
\ell_{\mathit{NLL}_{p}} = - \log p(f_a \mid f_b';\mathbf{d}_{ab}) - \log p(f_b \mid f_a';\mathbf{d}_{ba}).
\end{equation}

\noindent{Adaptive loss balancing:} to ensure stable training despite the different scales of our loss components, we dynamically balance them using:
\begin{equation}
\small
\ell_{\mathit{overall}} = \ell_{\mathit{NLL}_{p}} + \mathbf{w} \times \frac{\left|\ell_{\mathit{NLL}_{p}}\right|}{\left| \ell_{\mathit{NLL}_{m}} \right|} \cdot \ell_{\mathit{NLL}_{m}},
\end{equation}
where $\mathbf{w}$ is a predefined weight.
\section{Experiments}
\label{subsec:experiment}

\subsection{Dataset}
\label{subsec:dataset}
We evaluate our method on: CAHomo \cite{CA-Unsupervised2020} and GHOF \cite{gyroflow+}. We train on CAHomo (460K training pairs, 4.2K test pairs across regular, low-texture, low-light, and foreground scenes) with additional generated samples \cite{li2024dmhomo}, and conduct zero-shot testing on GHOF (256 test pairs in Regular, Foggy, Low-light, Rainy, and Snowy conditions).

\begin{figure}
    \centering
    \includegraphics[width=1\linewidth]{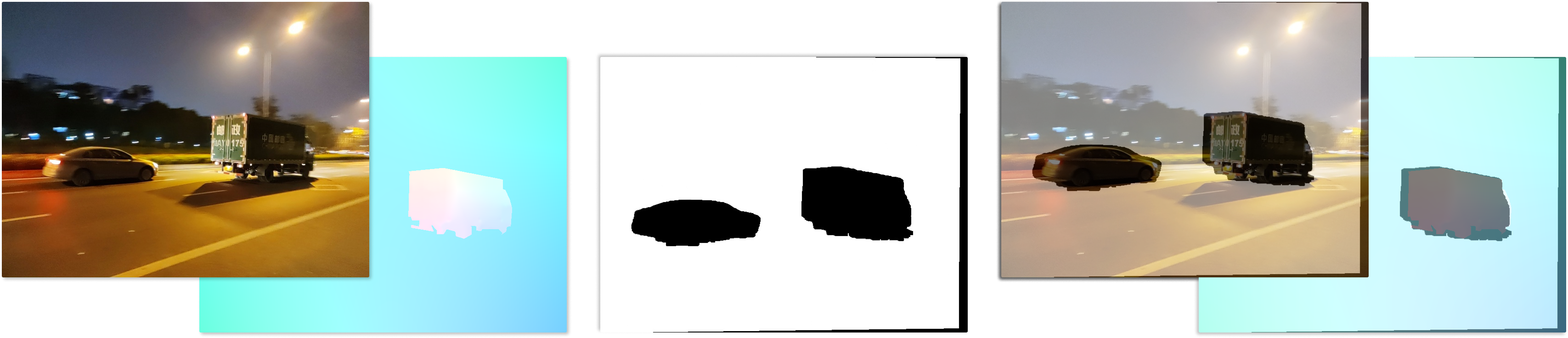}
    \vspace{-1em}
    \caption{Using the GHOF benchmark, we leverage SAM to generate semantic maps and manually identify dynamic objects such as cars and people, producing corresponding masks. These masks are then dilated to encompass occlusion regions, reducing ill-posed artifacts. Finally, the masks are applied to both the images and optical flow, isolating camera-induced motion and constructing a camera-motion-only dataset.}
    \label{fig:GHOF-Mask}
    \vspace{-1em}
\end{figure}

\noindent\textbf{GHOF-Cam Benchmark.} To isolate camera motion from dynamic scene elements, we propose a camera-motion-specific benchmark derived from GHOF. We employ the Segment Anything Model (SAM)~\cite{kirillov2023segment} to generate semantic maps, from which we identify dynamic objects (e.g., cars, people) and create corresponding masks. These masks are dilated to encompass occlusion regions, mitigating ill-posed artifacts at object boundaries. For edge occlusions not detected by semantic segmentation, we mainly utilize the ground-truth homography to identify black edge regions as additional masks. The combined masks are then applied to both input images and ground-truth optical flow, resulting in a benchmark that exclusively captures camera-induced motion, as illustrated in Fig.~\ref{fig:GHOF-Mask}.


\begin{table}[t]
  \begin{center}
  \resizebox{1\linewidth}{!}{
  \begin{tabular}{r|l|lllllll}
  \toprule
  & Methods & \multicolumn{1}{c}{AVG} & \multicolumn{1}{c}{RE} & \multicolumn{1}{c}{LT} & \multicolumn{1}{c}{LL} & \multicolumn{1}{c}{SF} & \multicolumn{1}{c}{LF} \\
  \midrule
  1) & $\mathcal{I}_{3\times3}$ & 6.70  & 7.75  & 7.65  & 7.21  & 7.53  & 3.39  \\
  \midrule
  2) & SIFT\cite{sift} + RANSAC\cite{ransac} & 1.41  & 0.30  & 1.34  & 4.03  & 0.81  & 0.57  \\
  3) & SIFT\cite{sift} + MAGSAC\cite{magsac} & 1.34  & 0.31  & 1.72  & 3.39  & 0.80  & 0.47  \\
  4) & ORB\cite{orb} + RANSAC\cite{ransac} & 1.48  & 0.85  & 2.59  & 1.67  & 1.10  & 1.24  \\
  5) & ORB\cite{orb} + MAGSAC\cite{magsac} & 1.69  & 0.97  & 3.34  & 1.58  & 1.15  & 1.40  \\
  6) & SPSG\cite{superpoint,superglue} + RANSAC\cite{ransac} & 0.71  & 0.41  & 0.87  & 0.72  & 0.80  & 0.75  \\
  7) & SPSG\cite{superpoint,superglue} + MAGSAC\cite{magsac} & 0.63  & 0.36  & 0.79  & 0.70  & 0.71  & 0.70  \\
  8) & LoFTR\cite{loftr} + RANSAC\cite{ransac} & 1.44  & 0.56  & 2.70  & 1.36  & 1.05  & 1.52  \\
  9) & LoFTR\cite{loftr} + MAGSAC\cite{magsac} & 1.39  & 0.55  & 2.57  & 1.33  & 1.05  & 1.41  \\
  \midrule

  10) & DHN\cite{supervised2016} & 2.87  & 1.51  & 4.48  & 2.76  & 2.62  & 3.00  \\
  11) & LocalTrans\cite{Crossresolution-supervised2021} & 4.21  & 4.09  & 4.84  & 4.55  & 5.30  & 2.25  \\
  12) & IHN\cite{Iterative-supervised2022} & 4.67  & 4.85  & 5.54  & 5.10  & 5.04  & 2.84  \\

  13) & {$\textrm{RealSH}$\cite{jiang2023supervised}} & {0.34 } & {0.22 } & {0.35 } & {{0.44} } & {0.42 } & {0.29 }\\
  14) & {$\textrm{DMHomo}$\cite{li2024dmhomo}} & {0.31 } & {0.19 } & {0.33 } & {0.40 } & {0.38 } & {0.28 } \\
  \midrule
    
  15) & CAHomo\cite{CA-Unsupervised2020} & 0.88  & 0.73  & 1.01  & 1.03  & 0.92  & 0.70  \\
  16) & BasesHomo\cite{BasesHomo2021} & 0.50  & 0.29  & 0.54  & 0.65  & 0.61  & 0.41  \\
  17) & HomoGAN\cite{HomoGAN2022} & {0.39 } & {0.22 } & {0.41 } & 0.57  & {0.44 } & {0.31 }\\
  \midrule
  18) & Ours$^{\star}$ & {0.33 } & {0.20 } & {0.33 } & {0.41 } & {0.39 } & {0.30 } \\
  19) & Ours & {0.32 } & {0.19 } & {0.32 } & {0.39 } & {0.39 } & {0.31 } \\
   \bottomrule
   \end{tabular}}
\end{center}
\vspace{-1.5em}
\caption{The benchmark consists of 5 distinct scenarios, namely regular (RE), low-texture (LT), low-light (LL), small foreground (SF), and large foreground (LF). The point matching errors (PME) on the test set of CAHomo~\cite{CA-Unsupervised2020} are presented.}
\vspace{-1em}
\label{tab:Compare_CAunsup}
\end{table}

\begin{table}[t]
  \begin{center}
  \resizebox{1\linewidth}{!}{
  \begin{tabular}{rl|ccccccc}
  \toprule
  1) & Methods & AVG & RE & FOG & LL & RAIN & SNOW \\
    \midrule
    2) & $\mathcal{I}_{3\times3}$ &5.22 &3.65 & 6.69 & 5.88  & 4.90  & 4.96  \\
    \midrule
    3) & SIFT\cite{sift} &2.82 &0.60 & 2.43 & 7.09  & 0.61  & 3.37  \\
    4) & SPSG\cite{superpoint,superglue} &3.07 &3.99 & 1.57 & 6.88  &0.79  & 2.16  \\
    5) & CAHomo\cite{CA-Unsupervised2020} &2.81 &2.02 & 2.03 & 4.56  & 2.84  & 2.61  \\
    \midrule
    6) & BasesHomo\cite{BasesHomo2021} &1.74 &1.39 & 0.97 & 4.12  & 0.66  & 1.58  \\
    7) & Meshflow\cite{LiuTYSZ16} &2.15 &1.09&2.21&5.57&0.44 &1.69 \\
    8) & HM\_Mix\cite{grundmann2012calibration} &4.35 &1.02 & 4.03 & 8.75  & 1.53  & 6.42  \\
    9) & RANSAC-F\cite{shen2020ransac} &3.26 &2.81 &3.14  &5.12  & 2.21  &3.04   \\
    \midrule
    10) & Ours &1.10 &1.08 & 0.74 & 2.15  & 0.46  & 1.05  \\
  \bottomrule
  \end{tabular}}
  \end{center}
  \vspace{-1.5em}
  \caption{To evaluate generalizability, we compute the end point errors (EPE) of pre-trained models from Table~\ref{tab:Compare_CAunsup} on our proposed GHOF-Cam benchmark.}
  \label{tab:GHOF-Cam-Quan}
  \vspace{-1em}
\end{table}%

\begin{table}[t]
  \begin{center}
  \resizebox{1\linewidth}{!}{
  \begin{tabular}{rl|ccccccc}
  \toprule
  1) & Methods & AVG & RE & FOG & LL & RAIN & SNOW \\
    \midrule
    2) & $\mathcal{I}_{3\times3}$ & 6.33  & 4.94  & 7.24  & 8.09  & 5.48  & 5.89  \\
      \midrule
    3) & SIFT\cite{sift} & 4.80 & {0.59}  & 4.47  & 12.10 & 0.62  & 6.20   \\
    4) & SPSG\cite{superpoint,superglue} & 4.47 & 3.54  & 2.21  & 10.66 & 0.83  & 5.10   \\
    
    \midrule
    5) & DHN\cite{supervised2016} & 6.61  & 6.04  & 6.02  & 7.68  & 6.99  & 6.32  \\
    6) & LocalTrans\cite{Crossresolution-supervised2021} & 5.72 & 4.06 & 6.49 & 5.95 & 5.78 & 6.34 \\
    7) & IHN\cite{Iterative-supervised2022} & 8.17 & 7.10 & 8.71 & 9.34 & 6.57 & 9.13 \\
    8) & RealSH\cite{jiang2023supervised} & {1.72} & 1.60 & 0.88 & 4.42 & {0.43} & {1.28} \\
    9) & DMHomo\cite{li2024dmhomo} & 1.75 & {0.64} & 0.85 & 4.16 & 0.39 & 2.74 \\
    \midrule
    10) & CAHomo\cite{CA-Unsupervised2020} & 3.87  & 4.10  & 3.84 & 6.99  & 1.27  & 3.17  \\
    11) & BasesHomo\cite{BasesHomo2021} & 2.28  & 2.02  & 1.43  & 4.90  & 0.78  & 2.29  \\
    12) & HomoGAN\cite{HomoGAN2022} & 1.95 & 1.73 & {0.60} & {3.95} & 0.47 & 3.02 \\
    \midrule
    13) & Ours & {1.23} & 1.15 & {0.96} & {2.69} & {0.40} & {0.93} \\
  \bottomrule
  \end{tabular}}
  \end{center}
  \vspace{-1.5em}
  \caption{To evaluate generalizability, we compute the point matching errors (PME) on the GHOF~\cite{gyroflow+} test set using pre-trained models from Table~\ref{tab:Compare_CAunsup}.}
  \label{tab:Compare_GHOF}
  \vspace{-1em}
\end{table}%

\begin{table*}[ht]
\centering
\resizebox{\linewidth}{!}{%
\begin{tabular}{l|ccc|ccc|ccc|ccc|ccc|ccc}
\toprule
\multirow{2}{*}{Method} & 
\multicolumn{3}{c|}{AVG} & 
\multicolumn{3}{c|}{RE} & 
\multicolumn{3}{c|}{FOG} & 
\multicolumn{3}{c|}{DARK} & 
\multicolumn{3}{c|}{RAIN} & 
\multicolumn{3}{c}{SNOW} \\
\cmidrule(lr){2-4} \cmidrule(lr){5-7} \cmidrule(lr){8-10} \cmidrule(lr){11-13} \cmidrule(lr){14-16} \cmidrule(lr){17-19}
 & PSNR$\uparrow$ & SSIM$\uparrow$ & LPIPS$\downarrow$ & PSNR$\uparrow$ & SSIM$\uparrow$ & LPIPS$\downarrow$ & PSNR$\uparrow$ & SSIM$\uparrow$ & LPIPS$\downarrow$ & PSNR$\uparrow$ & SSIM$\uparrow$ & LPIPS$\downarrow$ & PSNR$\uparrow$ & SSIM$\uparrow$ & LPIPS$\downarrow$ & PSNR$\uparrow$ & SSIM$\uparrow$ & LPIPS$\downarrow$ \\
\midrule
$\mathcal{I}_{3\times3}$ & 24.05 & 0.7403 & 0.0836 & 21.06 & 0.6900 & 0.0750 & 26.57 & 0.7711 & 0.0821 & 25.70 & 0.8506 & 0.0785 & 21.53 & 0.5335 & 0.1411 & 25.37 & 0.8562 & 0.0412 \\
GT-Homo & 32.78 & 0.9187 & 0.0570 & 28.39 & 0.8697 & 0.0549 & 35.23 & 0.9508 & 0.0492 & 31.88 & 0.9405 & 0.0575 & 30.11 & 0.8511 & 0.1033 & 38.31 & 0.9814 & 0.0199 \\
\midrule
SIFT & 28.44 & 0.9074 & 0.0781 & 29.23 & 0.9148 & 0.0545 & 29.42 & 0.9016 & 0.0768 & 27.37 & 0.9074 & 0.0982 & 30.00 & 0.8632 & 0.1055 & 26.16 & 0.9497 & 0.0558 \\
SPSG & 28.01 & 0.8697 & 0.0796 & 21.83 & 0.7593 & 0.0886 & 30.88 & 0.9049 & 0.0645 & 27.60 & 0.9019 & 0.0966 & 28.86 & 0.8270 & 0.1103 & 30.88 & 0.9556 & 0.0379 \\
CAHomo & 25.29 & 0.7837 & 0.0841 & 22.67 & 0.7341 & 0.0805 & 27.51 & 0.8048 & 0.0751 & 26.12 & 0.8743 & 0.0846 & 22.95 & 0.6130 & 0.1420 & 27.20 & 0.8924 & 0.0384 \\
\midrule
BasesHomo & 29.61 & 0.9026 & 0.0672 & 25.08 & 0.8522 & 0.0666 & 31.06 & 0.9170 & 0.0627 & 30.05 & 0.9303 & 0.0702 & 29.58 & 0.8512 & 0.1071 & 32.30 & 0.9622 & 0.0292 \\
MeshFlow & 29.91 & 0.9239 & 0.0688 & 28.57 & 0.9216 & 0.0576 & 28.68 & 0.9280 & 0.0742 & 29.41 & 0.9254 & 0.0774 & 30.68 & 0.8747 & 0.1049 & 32.23 & 0.9700 & 0.0298 \\
HM\_Mix & 25.77 & 0.8896 & 0.0882 & 26.09 & 0.8721 & 0.0596 & 26.56 & 0.8753 & 0.0882 & 26.43 & 0.9037 & 0.1002 & 28.20 & 0.8672 & 0.1107 & 21.58 & 0.9296 & 0.0820 \\
RANSAC-F & 26.04 & 0.8348 & 0.0890 & 26.09 & 0.8812 & 0.0665 & 29.22 & 0.8944 & 0.0801 & 27.29 & 0.9031 & 0.0923 & 21.68 & 0.5585 & 0.1495 & 25.90 & 0.9371 & 0.0566 \\
\midrule
Ours & 32.09 & 0.9142 & 0.0575 & 27.08 & 0.8615 & 0.0558 & 34.17 & 0.9371 & 0.0512 & 32.36 & 0.9421 & 0.0565 & 30.52 & 0.8608 & 0.1021 & 36.35 & 0.9692 & 0.0218 \\
\bottomrule
\end{tabular}
}
\vspace{-1em}
\caption{Quantitative comparison of different methods on various environmental conditions. We present three key perception metrics: PSNR (higher is better), SSIM (higher is better), and LPIPS (lower is better). Our method consistently outperforms existing approaches across all scenarios and metrics.}
\label{tab:ghof-cam}
\vspace{-1.5em}
\end{table*}

\subsection{Comparison with Existing Methods}
\label{subsec:compare_experiments}
We assess CamFlow against a comprehensive set of methods across three primary categories. The first category encompasses feature-based methods: SIFT \cite{sift}, ORB \cite{orb}, SuperPoint\cite{superpoint} with SuperGlue (SPSG) \cite{superglue}, and LoFTR \cite{loftr}, each evaluated with two outlier rejection techniques: RANSAC \cite{ransac} and MAGSAC \cite{magsac}. The second category includes supervised learning approaches: DHN \cite{supervised2016}, LocalTrans \cite{Crossresolution-supervised2021}, IHN \cite{Iterative-supervised2022}, RealSH \cite{jiang2023supervised}, and DMHomo \cite{li2024dmhomo}. The third category comprises unsupervised methods: CAHomo \cite{CA-Unsupervised2020}, BasesHomo \cite{BasesHomo2021}, and HomoGAN \cite{HomoGAN2022}.

For multi-plane camera motion modeling, we compare against both traditional approaches (MeshFlow \cite{LiuTYSZ16}, Homography Mixture \cite{grundmann2012calibration}, RANSAC-Flow \cite{shen2020ransac}) and unsupervised deep methods (BasesMesh \cite{liu2023unsupervised} and MeshHomoGAN \cite{liu2024unsupervised}). Regarding pre-training, DHN, LocalTrans and IHN use the MS-COCO dataset \cite{MSCOCO}, while other deep learning methods are pre-trained on CAHomo. Additional qualitative results and visual comparisons are available on: \url{https://lhaippp.github.io/CamFlow/}.

\subsubsection{Quantitative Comparison.}
\label{subsec:quanti_compare}


\begin{figure*}[th]
\begin{center}
  \includegraphics[width=1\linewidth]{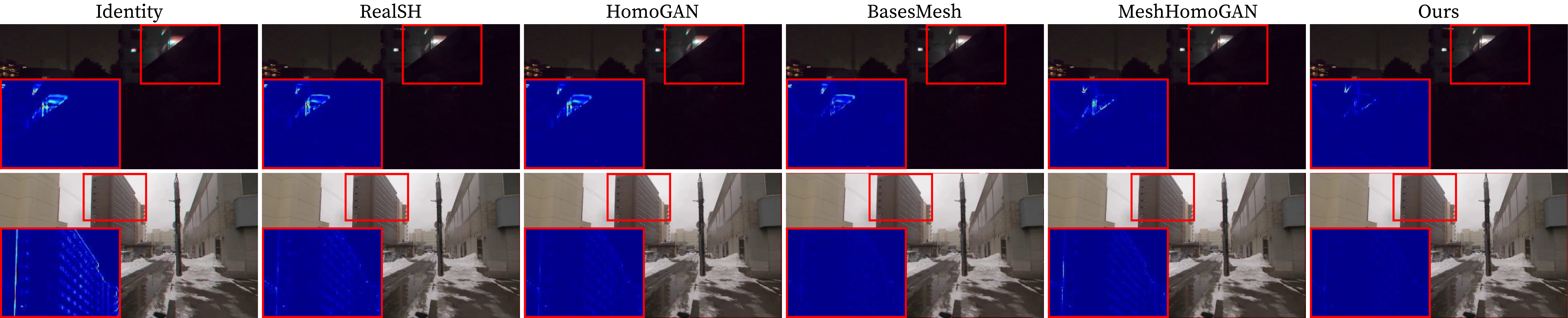}
\end{center}
\vspace{-1.5em}
  \caption{Qualitative results of CamFlow and methods from each category (i.e., supervised, unsupervised, and multi-homography) on the CAHomo testset~\cite{CA-Unsupervised2020}. The images are generated by superimposing the warped source images on the target image. Error-prone regions are highlighted with red boxes, which are further converted into alignment heatmaps for better distinction when zoomed in.}
    \vspace{-1em}
\label{fig:cahomo_qualitative}
\end{figure*}

\begin{figure*}[th]
\begin{center}
  \includegraphics[width=1\linewidth]{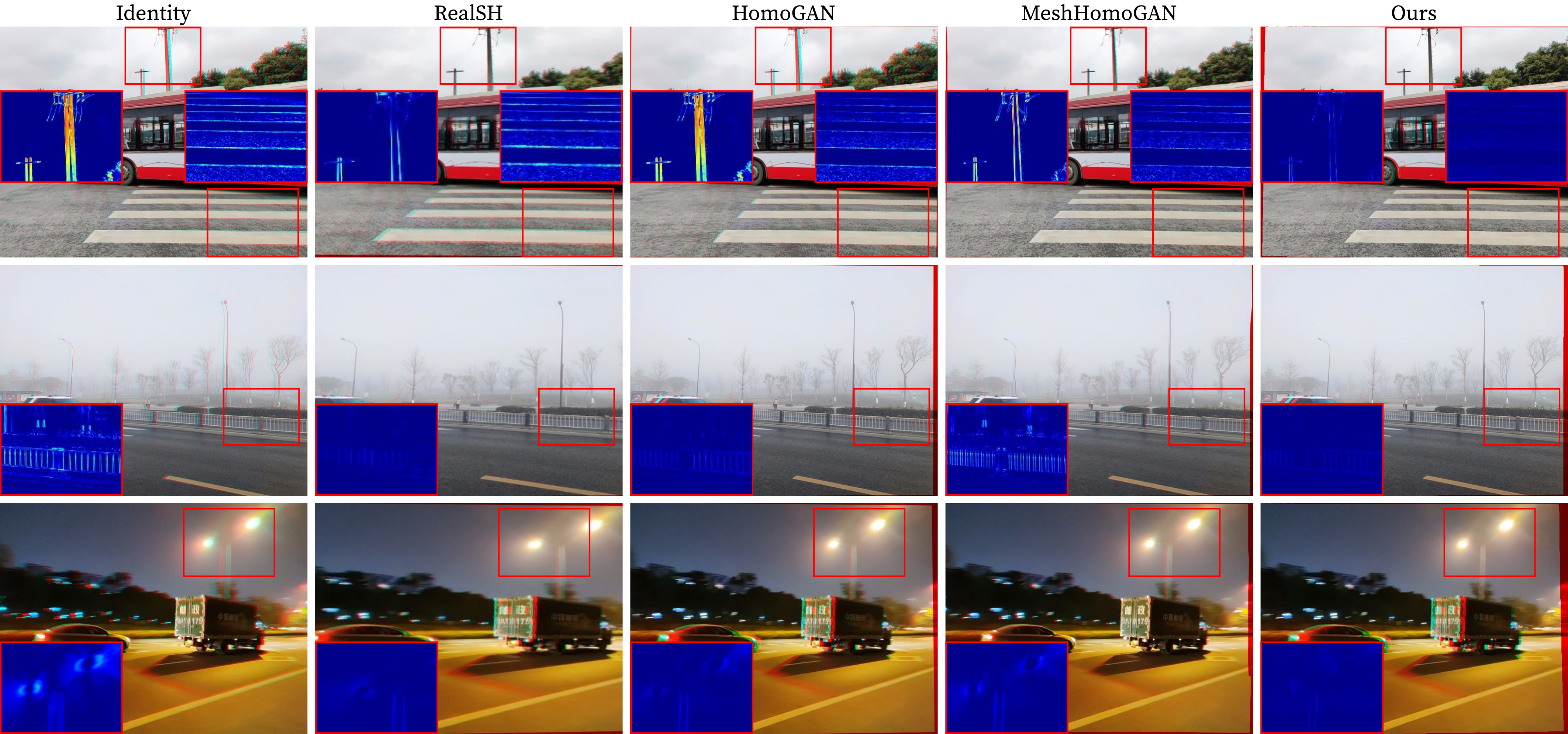}
\end{center}
\vspace{-1.5em}
  \caption{Qualitative results of our method and the best-performing generalizable methods from learning-based categories (i.e., supervised, unsupervised, and multi-homography) on the GHOF testset~\cite{gyroflow+}. The examples are arranged from top to bottom and include RE, Fog, and Dark. The images are generated by superimposing the warped source images on the target image. Error-prone regions are highlighted with red boxes, which are further converted into alignment heatmaps for better distinction when zoomed in.}
\label{fig:ghof_qualitative}
\vspace{-1em}
\end{figure*}

\noindent\textbf{Sparse camera motion.} We evaluate using the points matching error (PME), which measures the average geometric distance between transformed source points and their corresponding ground-truth target points. Table~\ref{tab:Compare_CAunsup} presents the quantitative performance of our method alongside various homography estimation approaches on the CAHomo test set, categorized as: feature-based methods (rows 2-9), supervised approaches (rows 10-14), and unsupervised ones (rows 15-17). The $\mathcal{I}_{3\times3}$ baseline (row 1) represents the distance between point pairs without transformation.

Our CamFlow method achieves superior results across multiple categories, outperforming the leading unsupervised method, HomoGAN, by reducing PME by 17.95\% (from 0.39 to 0.32). While feature-based methods like SIFT+RANSAC excel in regular (RE) scenes with abundant texture and keypoints, CamFlow surpasses them with a 36.67\% improvement (reducing error by 0.11). To balance benchmark performance with generalization capability, we include Ours$^{\star}$ (row 18), which represents an early-stopped training model that delivers better generalizability across datasets, even though it slightly underperforms our fully-trained model (row 19) on the CAHomo benchmark.

In small foreground (SF) and large foreground (LF) scenarios, where dynamic objects disrupt camera motion estimation, our probabilistic loss and confidence masking yield lower PMEs compared to methods like CAHomo and HomoGAN, which employ explicit outlier rejection masks for robustness. In low-texture (LT) and low-light (LL) conditions, learning-based approaches generally exhibit superior resilience due to their keypoint-free strategies, particularly among unsupervised techniques. However, these scenes often contain homogeneous regions~\cite{GLU-Net}, occupying large image areas and reducing photometric differences, which limit unsupervised methods’ effectiveness. By contrast, CamFlow excels in LT and LL.

\noindent\textbf{Generalization experiment.} Evaluating cross-dataset generalization presents a significant challenge in motion estimation, particularly for applications where real-world video often differs substantially from training data. We assess CamFlow's generalization capabilities in two ways: (1) for dense camera motion, we evaluate against traditional single-plane, multi-planar, and deep learning approaches using our proposed GHOF-Cam benchmark (Table \ref{tab:GHOF-Cam-Quan} and Table \ref{tab:ghof-cam}); and (2) for sparse camera motion, we compare against traditional and homography-based methods on the original GHOF benchmark \cite{gyroflow+} (Table \ref{tab:Compare_GHOF}).

Table~\ref{tab:GHOF-Cam-Quan} presents End Point Error (EPE) results on the GHOF-Cam benchmark, which represents ground-truth camera motion. We evaluate against homography-based methods (rows 3-5) and non-single-plane approaches (rows 6-9). As illustrated in Fig.~\ref{fig:Motion-Basis}, we classify BasesHomo~\cite{BasesHomo2021} as a multi-plane rather than single-homography method. The results demonstrate that CamFlow outperforms competing methods across nearly all categories, showcasing exceptional zero-shot capability in capturing non-linear camera motion patterns.
Additionally, Table~\ref{tab:ghof-cam} presents perceptual quality metrics (PSNR, SSIM, and LPIPS) across various conditions. We also include ground-truth homography results for reference (GT-Homo). Our method, CamFlow, achieves state-of-the-art performance across multiple categories, consistently outperforming all baselines, particularly in challenging conditions such as snowy scenes. Notably, CamFlow approaches the performance of ground-truth homography, with only a 0.69 dB difference in PSNR, 0.0045 in SSIM and 0.0005 in LPIPS, while significantly surpassing the second-best method, MeshFlow. These results support CamFlow's potential for applications requiring high-quality visual alignment.

Table~\ref{tab:Compare_GHOF} presents compelling evidence of CamFlow's generalization capability on the GHOF benchmark. Our method achieves the lowest average PME (1.23), representing a 28.5\% improvement over the previous best supervised method (RealSH at 1.72) and a 36.9\% improvement over the leading unsupervised approach (HomoGAN at 1.95). This performance advantage extends across all environmental conditions, with particularly significant gains in challenging scenarios. In low-light (LL) conditions, CamFlow reduces error by 39.1\% compared to RealSH (from 4.42 to 2.69), while in snowy (SNOW) environments, it achieves a remarkable 69.2\% error reduction compared to HomoGAN (from 3.02 to 0.93). These results confirm that CamFlow enables robust zero-shot transfer to unseen datasets, even under diverse and challenging environmental conditions.

\subsection{Qualitative Comparison}
\label{subsec:qualit_compare}

Fig.~\ref{fig:cahomo_qualitative} and \ref{fig:ghof_qualitative} present qualitative results of CamFlow alongside competing methods on the CAHomo and GHOF test sets. Dynamic visualizations are available on our project page.
For visualization, we employ red-blue ghosting and alignment heat maps following \cite{LBHomo}. After transforming the source frame using estimated camera flow, misaligned regions appear as red and blue ghosting. We highlight specific areas with red boxes and generate alignment heat maps where brighter regions indicate poorer alignment.

In both figures, the ``Identity" column shows source and target images overlaid without warping. For CAHomo (Fig.~\ref{fig:cahomo_qualitative}), we focus on evaluating background camera motion modeling capabilities. We selected challenging test cases featuring extremely low-light conditions (first row) and sophisticated motion patterns (parallax and depth variation), comparing against leading methods from three categories: supervised (RealSH), unsupervised (HomoGAN), and deep meshflow (BasesMesh, MeshHomoGAN). CamFlow achieves superior alignment in background regions.

The GHOF benchmark (Fig.~\ref{fig:ghof_qualitative}) highlights CamFlow's advantages in handling complex camera motion and its zero-shot generalization capabilities. The first row shows a scene with parallax and foreground motion that challenges homography-based methods and reduces meshflow accuracy. The second row demonstrates CamFlow's generalization to foggy conditions unseen in training data, where depth variation complicates motion estimation. The last row presents an extreme case with motion blur, large parallax, and dynamic objects. In all scenarios, CamFlow delivers robust results where competing methods struggle, confirming its effectiveness in modeling complex camera motion patterns across diverse environmental conditions.

\begin{figure}[t]
    \centering
    \noindent \includegraphics[width=\linewidth]{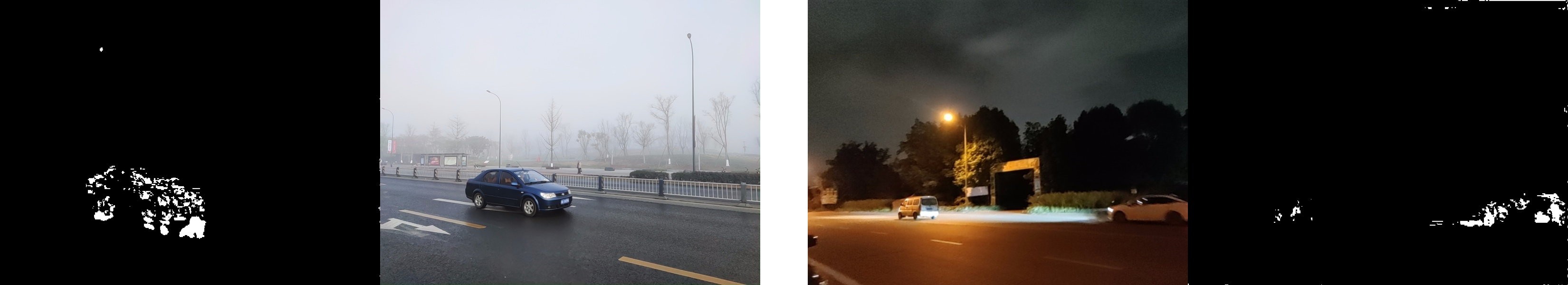}
    \caption{Uncertainty masks and corresponding original images. The brighter the mask, the higher the uncertainty. The masks effectively highlight dynamic objects, indicating regions where camera motion estimation is less reliable.}
    \label{fig:mask}
    \vspace{-1.5em}
\end{figure}

\noindent\textbf{Foreground Masks.} Fig.~\ref{fig:mask} illustrates the uncertainty masks. We observe that the network effectively identifies dynamic objects and assigns higher uncertainty to regions containing these objects. This design helps focus the learning process on areas where motion is most relevant, while reducing noise from dynamic object regions, thereby enhancing the accuracy of camera motion estimation.



\subsection{Ablation Studies}
\label{sec:abla_study}


\subsubsection{Motion Basis}
\vspace{-0.5em}
\begin{table}[h]
    \begin{center}
    \resizebox{1\linewidth}{!}{
    \begin{tabular}{r|ccc|cc}
    \toprule
    & CAHomo & GHOF & GHOF-Cam & Params & Inference Time \\
    \midrule
    8 Bases & 0.37  & 1.68  & 1.45 &2.658M & 76.42ms \\
    12 Bases & 0.36  & 1.54  & 1.23 &2.658M &75.38ms \\
    \midrule
    24 Bases & 0.33  & 1.23 & 1.10 &2.660M &79.63ms \\
    200 Bases  & 0.33  & 1.27  & 1.07 &2.677M &99.28ms \\
    \bottomrule
    \end{tabular}}
    \end{center}
    \vspace{-1.5em}
    \caption{Performance comparison under different numbers of motion bases on three benchmarks. Results demonstrate the effectiveness of combining physical and stochastic motion bases, with 24 bases providing optimal balance between accuracy and computational efficiency.}
    \label{tab:ablat1}
    \vspace{-1em}
\end{table}

In Table~\ref{tab:ablat1}, we evaluate with varying numbers of motion bases across three benchmarks: CAHomo (trained) and GHOF/GHOF-Cam (zero-shot). The average results demonstrate that: 1) Increasing physical bases from 8 to 12 improves performance across all benchmarks; 2) Introducing additional hybrid bases (24 total) yields further enhancements, particularly for generalization (GHOF and GHOF-Cam). Notably, while expanding to 200 bases provides marginal improvements on some benchmarks, it increases inference time by 24.7\%. We therefore adopt 24 bases in our final model as the optimal balance between accuracy and computational efficiency.


\subsubsection{Hybrid Probabilistic Loss}
\vspace{-0.5em}
\begin{table}[h]
    \begin{center}
    \resizebox{0.8\linewidth}{!}{
    \begin{tabular}{cc|ccc}
    \toprule
      $\ell_{NLL_m}$ & $\ell_{NLL_p}$ & CAHomo & GHOF & GHOF-Cam \\
    \midrule
    
     \Checkmark &  & 0.41  & 2.21  & 2.13  \\
      & \Checkmark & 0.36  & 1.58  & 1.42  \\
     \Checkmark & \Checkmark & 0.33  & 1.23  & 1.10  \\

    \bottomrule
    \end{tabular}}
    \end{center}
    \vspace{-1.5em}
    \caption{Ablation study of different loss function combinations across three benchmarks. Results demonstrate that the hybrid approach combining motion loss and photometric loss achieves superior performance compared to individual loss components.}
    \vspace{-1em}
    \label{tab:abla2}
\end{table}
Table~\ref{tab:abla2} evaluates our probabilistic loss components across three benchmarks. Using only motion loss yields limited generalization performance, because pseudo motion labels provide approximate supervision. The photometric loss alone performs substantially better, particularly on zero-shot datasets, consistent with findings from prior unsupervised methods. However, our hybrid approach combining both losses achieves the best results across all benchmarks, with improvements on GHOF-Cam (22.5\% error reduction compared to photometric-only). We believe motion labels provide coarse guidance while photometric loss enables fine-grained refinement, resulting in more accurate and generalizable camera motion estimation.

\section{Conclusion}
CamFlow presents a novel motion representation for modeling 2D camera motion using a hybrid motion basis approach. We identify a fundamental issue: superposing homographies by simply adding flow fields introduces non-linear interactions, contradicting the assumption that homographies can be expressed as linear combinations of 8 basis flows.
By expanding the previous 8-dimensional motion basis into a higher-dimensional space with both physical and stochastic motion bases, CamFlow effectively captures complex, non-linear motion patterns. The proposed probabilistic loss function enhances training stability.
Our newly introduced GHOF-Cam benchmark demonstrates that CamFlow surpasses state-of-the-art homography and meshflow methods, showcasing superior robustness and generalization. We hope this work opens new avenues for camera motion modeling in video processing applications.
Codebase, pre-trained models, and benchmark data are released at \url{https://lhaippp.github.io/CamFlow/}.

\vspace{0.5em}
\noindent\textbf{Acknowledgements.} 
This work was supported in part by the National Natural Science Foundation of China (NSFC) under grants No. 62372091 and No. 62031009, and in part by the Hainan Province Key R\&D Program under grant No. ZDYF2024(LALH)001.
We would also like to thank Hao Xu, Mingbo Hong, Hai Jiang, Xinglong Luo, Nianjin Ye, and the anonymous reviewers for their valuable discussions, constructive suggestions, and helpful feedback.

\clearpage
{
    \small
    \bibliographystyle{ieeenat_fullname}
    \bibliography{main}
}

\end{document}